\documentclass[10pt,twocolumn,letterpaper]{article}

\usepackage{cvpr}              % To produce the CAMERA-READY version
\usepackage{graphicx}
\usepackage{amsmath}
\usepackage{amssymb}
\usepackage{booktabs}

\usepackage[pagebackref,breaklinks,colorlinks]{hyperref}

\usepackage[capitalize]{cleveref}
\crefname{section}{Sec.}{Secs.}
\Crefname{section}{Section}{Sections}
\Crefname{table}{Table}{Tables}
\crefname{table}{Tab.}{Tabs.}

\def\cvprPaperID{3427} % *** Enter the CVPR Paper ID here
\def\confName{CVPR}
\def\confYear{2022}

\begin{document}

% \title{Structure Xfer: Leveraging Structure And Color Layers For Efficient Incremental Dynamic View Synthesis In Seconds}
\title{INV: Towards Streaming Incremental Neural Videos}
% \title{INV: Leveraging Natural Information Partitions in MLPs\\ to Stream Incremental Neural Videos}

% \author{Shengze Wang\\
% UNC Chapel Hill\\
% %Institution1 address\\
% {\tt\small shengzew@cs.unc.edu}
% \and

\author{Shengze Wang$^{1,2}$ \quad Alexey Supikov$^2$ \quad Joshua Ratcliff$^2$ \quad Henry Fuchs$^1$ \quad Ronald Azuma$^2$\\
$^1$UNC Chapel Hill \quad\quad\quad\quad\quad\quad\quad\quad\quad\quad\quad\quad\quad $^2$Intel Labs\quad\quad\quad\quad\quad\quad\quad\quad\quad\quad\quad\\
{\tt\small \{shengzew,fuchs\}@cs.unc.edu} \quad 
{\tt\small \{Alexei.Soupikov,joshua.j.ratcliff,ronald.t.azuma\}@intel.com}
}
% UNC Chapel Hill\\
% {\tt\small shengzew@cs.unc.edu}

\maketitle

%%%%%%%%% ABSTRACT
\begin{abstract}
Recent works in spatiotemporal radiance fields can produce photorealistic free-viewpoint videos. However, they are inherently unsuitable for interactive streaming scenarios (\eg video conferencing, telepresence) because have an inevitable lag even if the training is instantaneous. This is because these approaches consume videos and thus have to buffer chunks of frames (often seconds) before processing. 

In this work, we take a step towards interactive streaming via a frame-by-frame approach naturally free of lag. Conventional wisdom believes that per-frame NeRFs are impractical due to prohibitive training costs and storage. We break this belief by introducing Incremental Neural Videos (INV), a per-frame NeRF that is efficiently trained and streamable. We designed INV based on two insights: 
% Moreover, INV is a general framework that can be combined with more advanced models to achieve even better results faster. 

(1) Our main finding is that MLPs naturally partition themselves into Structure and Color Layers, which store structural and color/texture information respectively.

(2) We leverage this property to retain and improve upon knowledge from previous frames, thus amortizing training across frames and reducing redundant learning. 

As a result, with negligible changes to NeRF, INV can achieve good qualities ($>28.6db$) in 8min/frame. It can also outperform prior SOTA in $19\%$ less training time. Additionally, our Temporal Weight Compression reduces the per-frame size to 0.3MB/frame ($6.6\%$ of NeRF). More importantly, INV is free from buffer lag and is naturally fit for streaming. 
While this work does not achieve real-time training, it shows that incremental approaches like INV present new possibilities in interactive 3D streaming.
Moreover, our discovery of natural information partition leads to a better understanding and manipulation of MLPs. Code and dataset will be released soon.

\end{abstract}

%%%%%%%%% BODY TEXT
\section{Introduction}
\label{sec:intro}

\setlength{\belowcaptionskip}{-10pt}
\begin{figure}[t]
  \centering
   \includegraphics[width=0.45\textwidth]{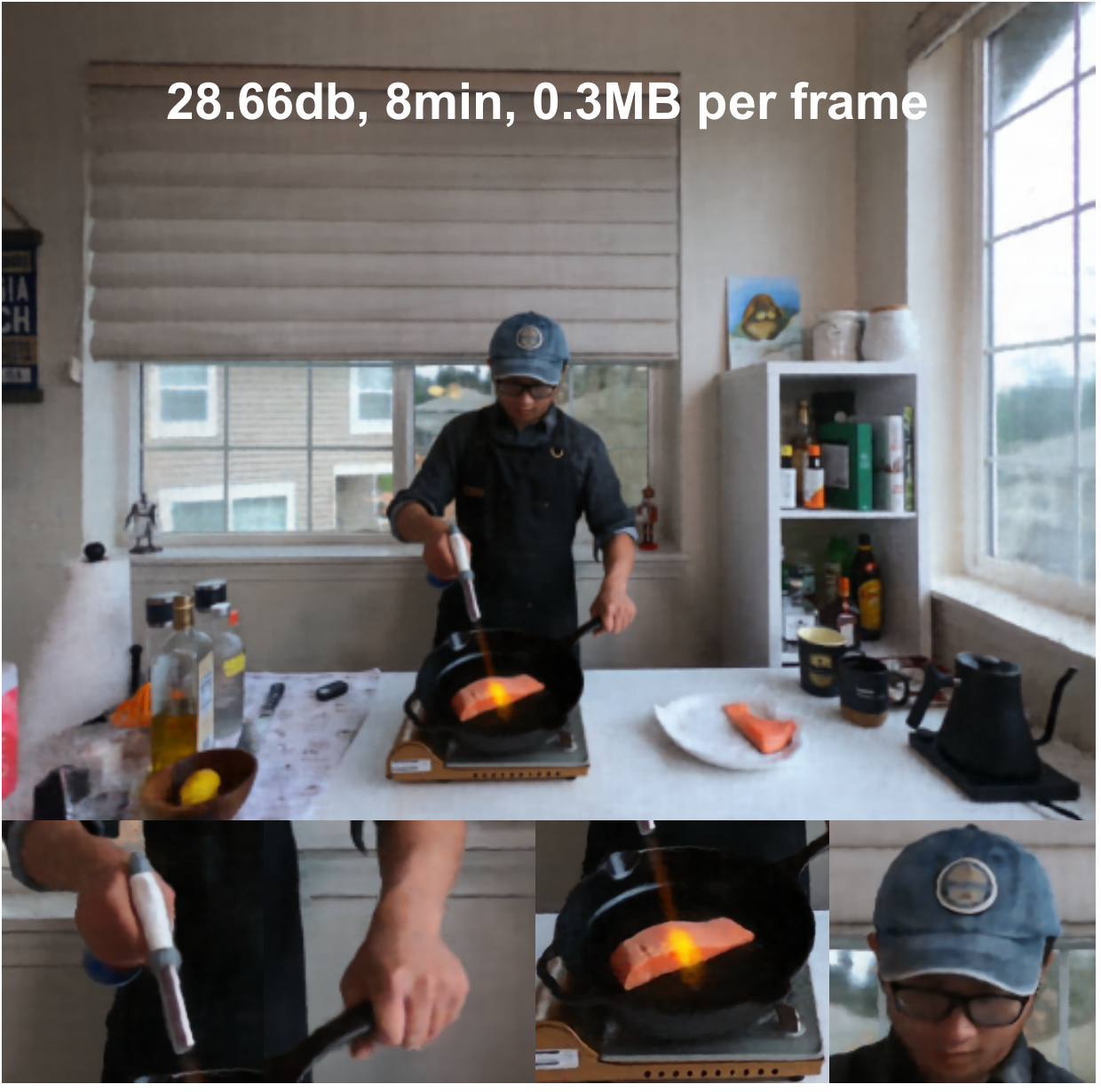}

   \caption{An example rendered novel view rendered by INV on \textbf{\textit{Plenoptic Video Datasets \cite{neural3dvideos}}}. INV uses the original NeRF as the rendering backbone and averages over 28.7db PSNR with only 0.3MB of storage and 8 minutes of training per frame.}
   \label{fig:teaser}
\end{figure}
\setlength{\belowcaptionskip}{0pt}

% we need to describe 3D live-streaming more at the be-
% ginning, as we get more compressed representation, and re-
% altime rendering, we could stream such representations to
% observers’ machines. And their machines can render in 3D.)
% (we aim that in the near future, we envision people can
% do 3D youtube videos with high-end GPUs.

Recent advances in photorealistic 3D video generation hint at an exciting future where moments can be captured and experienced in 3D. One application of peculiar potential is interactive 3D streaming. It is the basis of telepresence systems, which can democratize high-quality education via immersive remote classrooms, replace business traveling, and allow for face-to-face reunions between distant friends and families. In this work, we take a step towards this exciting future by introducing an efficient representation suitable for interactive 3D video streaming. 

Spatiotemporal NeRFs\cite{neural3dvideos,nsff,dynamicnerf,torf}) are a popular approach to synthesizing 3D videos due to their impressive photorealism. However, they suffer from an inherent lag because they consume videos and thus have to wait for chunks of frames (often seconds) before processing. Due to this buffer lag, these approaches are not suitable for interactive streaming scenarios like telepresence and immersive cloud gaming. In this work, we introduce Incremental Neural Videos (INV), a frame-by-frame approach that is naturally free of lag. While it is a common belief that per-frame NeRFs are impractical due to their prohibitive training costs and storage size, INV breaks this belief by showing the possibility of training each frame in minutes and maintaining a streamable size. INV achieves this with negligible changes to NeRF, and it is a general framework that can be combined with more optimized approaches.

Moreover, prior approaches mostly show results on short videos lasting several seconds. This is because they model the entire 3D video as a single spatiotemporal object, and thus are limited by small model capacities. While one can divide longer videos into chunks (as in Neural 3D Videos\cite{neural3dvideos}), it could months of GPU hours to generate a minute-long 3D video. The high training costs thus prevent the general public from using them.

We argue that the cause of such high training costs, in addition to the commonly discussed sampling bottleneck, can also be attributed to the learning of duplicated information. Prior works\cite{neural3dvideos,dynamicnerf,nsff,nerf} learn the mapping from input 4D points $[x,y,z,t]$ (or equivalent) to their radiance. This input formulation essentially views 3D points from different frames as different 4D points. As a result, even if the scene is static, the same 3D points would need to be re-learned when the time variable changes. 
In our work, we retain and improve upon previously learned information to avoid training each frame from scratch and to continually improve over time. Moreover, we discovered that MLPs naturally store structural information in earlier Structure Layers and color information in later Color Layers. Since the color/texture space is often consistent in many scenarios (\eg teleconferencing and live streaming), we significantly reduce storage by maintaining a constant color/texture space and learning only structural information.

Based on these observations, we propose Incremental Neural Videos (INV), a neural representation that can be efficiently learned and stored frame-by-frame. Our contributions can be summarized as follows:
\begin{itemize}
  \item Natural Partition of Structure and Color Layers: we found that MLPs naturally store structural in earlier layers and color/texture information  in later layers. We perform numerous experiments on both 2D and 3D videos to showcase this phenomenon. This discovery leads to a clearer understanding and more effective manipulation of MLPs.
  \item We leverage this phenomenon to design Incremental Neural Videos (INV), which is naturally free of lag and suitable for interactive streaming. INV consists of two types of sub-modules: (1) a shared color module that is shared across frames and encodes the color/texture of the scene, and (2) per-frame structure modules that encode the changing structures of the dynamic scene and are stored frame-by-frame.
  \item We propose Structure Transfer, an incremental training scheme that significantly accelerates training. With Structure Transfer, INV outperforms the state-of-the-art on per-frame quality metrics with less training. Moreover, INV can achieve good qualities in minutes.
  \item We propose Temporal Weight Compression to further compress the already concise representation. A compressed INV frame is only 0.3MB, $6.6\%$ of NeRF.
\end{itemize}

%-------------------------------------------------------------------------
\section{Related Works}
% \begin{figure*}
%   \centering
%   % \fbox{\rule{0pt}{2in} \rule{0.9\linewidth}{0pt}}
%    % \includegraphics[scale=0.45]{images/struct_swap.png}
%   % \includegraphics[scale=0.21]{images/color_struct_info.png}
%   \includegraphics[scale=0.29]{images/structure_swap_captioned.pdf}
%    \caption{\textbf{Structure and Color Layers:} We found that MLPs naturally store structural information in \textbf{earlier} Structure Layers and color information in \textbf{later} Color Layers. Empirically, we observe 1 Structure Layer for 2D MLPs and 3-7 layers for NeRFs.\\
%    \textbf{Left:} When an MLP is trained incrementally from frame A to B, the Color Layers remain similar among nearby frames. Thus, replacing A's Color Layers with B's does not change the structural content. This phenomenon can be observed in both 2D and 3D. Moreover, replacing structure layers induces continual and meaningful motion in 3D.\\
%    \textbf{Right:} Color scheme transferred between images by freezing Color Layers of image A while training Structure Layers on image B.
%    }
%    \label{fig:swap}
% \end{figure*}

   % \textbf{Bottom-Left:} In NeRF, replacing more and more levels of Structural Layers causes more and more changes to 3D structures.\\
   % \textbf{Right:} MLP A is trained to reconstruct image A, then A's Structure Layers are briefly fine-tuned on image B (with Color Layer frozen). The color scheme of A is often preserved at early stages of the fine-tuning.

\subsection{Static Novel View Synthesis}
\textbf{Layered Representations.} Existing methods differ in how they represent 3D scenes. Multiplane Images (MPI) \cite{szeliski1998stereo,zhou2018stereo,srinivasan2019pushing,single_view_mpi,MatryODShka,googlelightfield} are variable-viewpoint images that represent a scene as a set of fronto-parallel images, layered-meshes, or multilayer-spheres. However, these methods usually exhibit stack-of-card artifacts when viewed at a close distance or from steep angles.

\textbf{Neural Radiance Fields.} NeRF \cite{nerf} introduced a simple but powerful representation based on Multi-Layer Perceptrons (MLPs) and differentiable volumetric rendering. It achieved unprecedented photorealism while allowing free viewpoints rendering. The high-level idea is to use a neural network to model the radiance of 3D points in the target scene. Notably, positional encodings\cite{ffn,siren} play a vital role in improving the quality of the reconstructed radiance field. Many works (\eg \cite{nerf++,mipnerf,ibrnet,mvsnerf}) have been inspired by NeRF. However, static NeRF faced two main challenges: 

(1) Slow training and rendering. Multiple approaches (\eg \cite{autoint,instantNGP,plenoctrees,plenoxels,bakingnerf,fastnerf,dsnerf,tensorrf}) have been proposed to improve training and rendering speed, but long training time is still the main bottleneck of NeRF-based models. 

(2) Generalization to new scenes. Many works like \cite{pixelnerf,mvsnerf,ibrnet,SRF} have tried to achieve generalization by adding ConvNets or Transformers to NeRF, but it remains a challenge to learn generalizable neural radiance fields.

\textbf{Explicit 3D Representations.} One of the fundamental difficulties in view synthesis is to model the scene in 3D based on 2D images. To alleviate the burden from view synthesis algorithms, some works \cite{FVS,SVS,PBNRPO,ADOP,npbg} rely on preprocessed raw 3D geometry (\eg point clouds and meshes), obtained from multi-view stereo software such as COLMAP\cite{colmap}. Such approaches show fewer fog-like artifacts which are common in NeRF-based methods. Methods like \cite{FVS,SVS} also demonstrate good generalization to new scenes. Point cloud-based neural rendering approaches \cite{PBNRPO,ADOP,wiles2020synsin} show impressive sharpness and details in large scenes with thin structures. However, the aforementioned approaches rely on good input 3D models to render high-quality images.

\subsection{Dynamic Novel View Synthesis}
\textbf{Human-Specific Approaches.} Some recent works \cite{neuralbody,neuralactor,anerf,animatablenerf,neuralhumanperformer} focus on animating clothed humans only. They often use colored videos as inputs and leverage human body templates (\eg SMPL\cite{SMPL} and STAR\cite{STAR}) and deep textures. On the other hand, LookinGood\cite{lookingood} uses multiple RGBD cameras to reconstruct human mesh in real-time. They then use a neural network to render colored mesh into high-quality novel views. HVSNet \cite{hvsnet} uses monocular RGBD videos and renders from feature point clouds. Our work focuses on general scenes with complex human-object interactions. Therefore, approaches limited to human subjects would not apply to our target scenario.

\textbf{Dynamic Novel View Synthesis of General Scenes} Yoon et al.\cite{nvidiadynamicnvs} utilizes multi-view stereo and monocular depth estimators to estimate spatially and temporally coherent depth maps and motion fields. Their approach generates 3D videos without per-video optimization or prior knowledge of the scene. Many NeRF-based approaches\cite{videonerf,nsff,dynamicnerf,nonrigidnerf} encode dynamic scenes as spatio-temporal radiance fields. Many learn a radiance field and a motion field for each frame. The motion field is then used to establish 3D correspondences between frames, thus allowing for temporal consistency regularizations. Optical flow and monocular depth estimators are often used to guide the optimization to a good local optimum. T{\"o}RF\cite{torf} uses Time-of-Flight sensors to achieve better modeling of both static and dynamic scenes. Approaches like Nerfies\cite{nerfies}, HyperNeRF\cite{hypernerf}, and Neural 3D Videos (N3V)\cite{neural3dvideos} use latent codes to help model dynamic contents. N3V also shows that smart sampling strategies can accelerate training and improve rendering qualities. However, all of these approaches require hours of training per frame, leading to weeks of training for a short video lasting several seconds. Our work shows how to significantly accelerate training and reduce storage via a concise incremental representation that naturally extends to streaming applications.

\section{Method}
We discuss our work in 3 parts: (1) the Incremental Transfer (I.T.) training scheme that drastically accelerates the training. I.T. also leads to (2) our discovery of Structure and Color Layers in MLPs, and (3) the Incremental Neural Videos (INV) representation that leverages this discovery to drastically reduce storage size and increase stability.

\subsection{Transferring Information Across Frames}
\label{sec:IT}
Prior works \cite{neural3dvideos,nsff,videonerf,dynamicnerf,nonrigidnerf} can render impressively realistic 3D videos after hours of training per frame. The cause of such high training costs, aside from the well-explored sampling bottleneck, can also be attributed to the redundancy in training. Many works map 4D points $[x,y,z,t]$ (or equivalent) to $RGB\sigma$ values. This input formulation essentially views 3D points from different frames as different 4D points. Therefore, even if the scene is static, the same 3D points would need to be re-learned when the time variable changes. While loss terms enforce consistencies between mappings for different frames, duplicated mappings still consume model capacity and training resources. A similar argument also applies to dynamic points. 

\begin{figure*}
  \centering
   % \includegraphics[scale=0.45]{images/struct_swap.png}
  % \includegraphics[scale=0.21]{images/color_struct_info.png}
  \includegraphics[width=\textwidth]{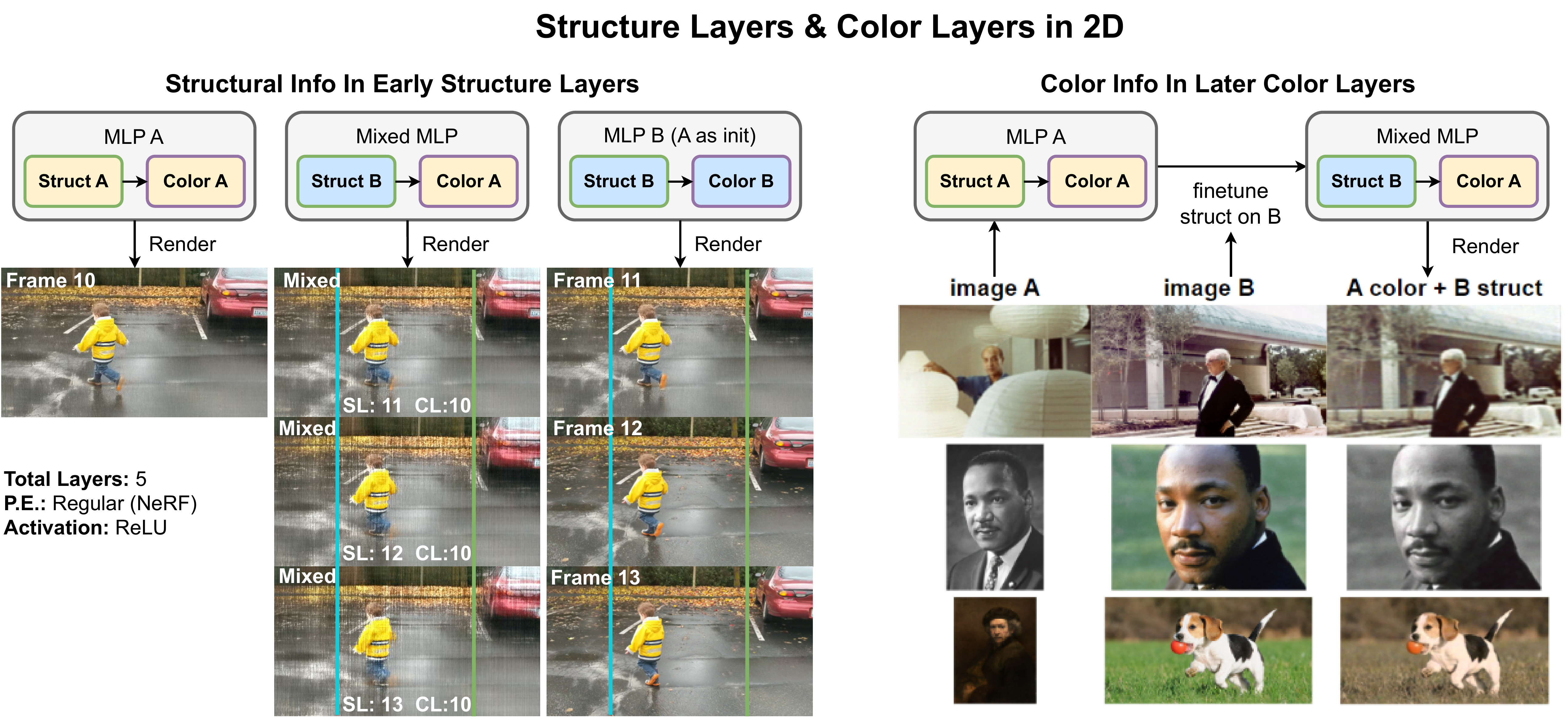}
   \caption{We found that MLPs naturally store structural and color information separately in \textbf{earlier} Structure Layers and \textbf{later} Color Layers. 
   \textbf{Structure Swap (Left):} When an MLP is trained \textbf{incrementally} from frame A to B, color information remains similar among nearby frames but structural information varies. As a result, replacing A's Structure Layers (Struct A) with B's Structure Layers (Struct B) would induce meaningful structural changes/movements \textbf{without training}. Empirically, we observe 1 Structure Layer for 2D MLPs.\\
   \textbf{Color Scheme transfer (Right):} One can often transfer color schemes between images by (1) training MLP A on image A, and then \\
   (2) finetuning A's SLs (with CLs frozen) on image B. During the early stages of finetuning, the color scheme of A is often preserved.
   }
   \label{fig:2d_swap}
\end{figure*}

\setlength{\belowcaptionskip}{-15pt}
\begin{figure}
  \centering
   \includegraphics[scale=0.44]{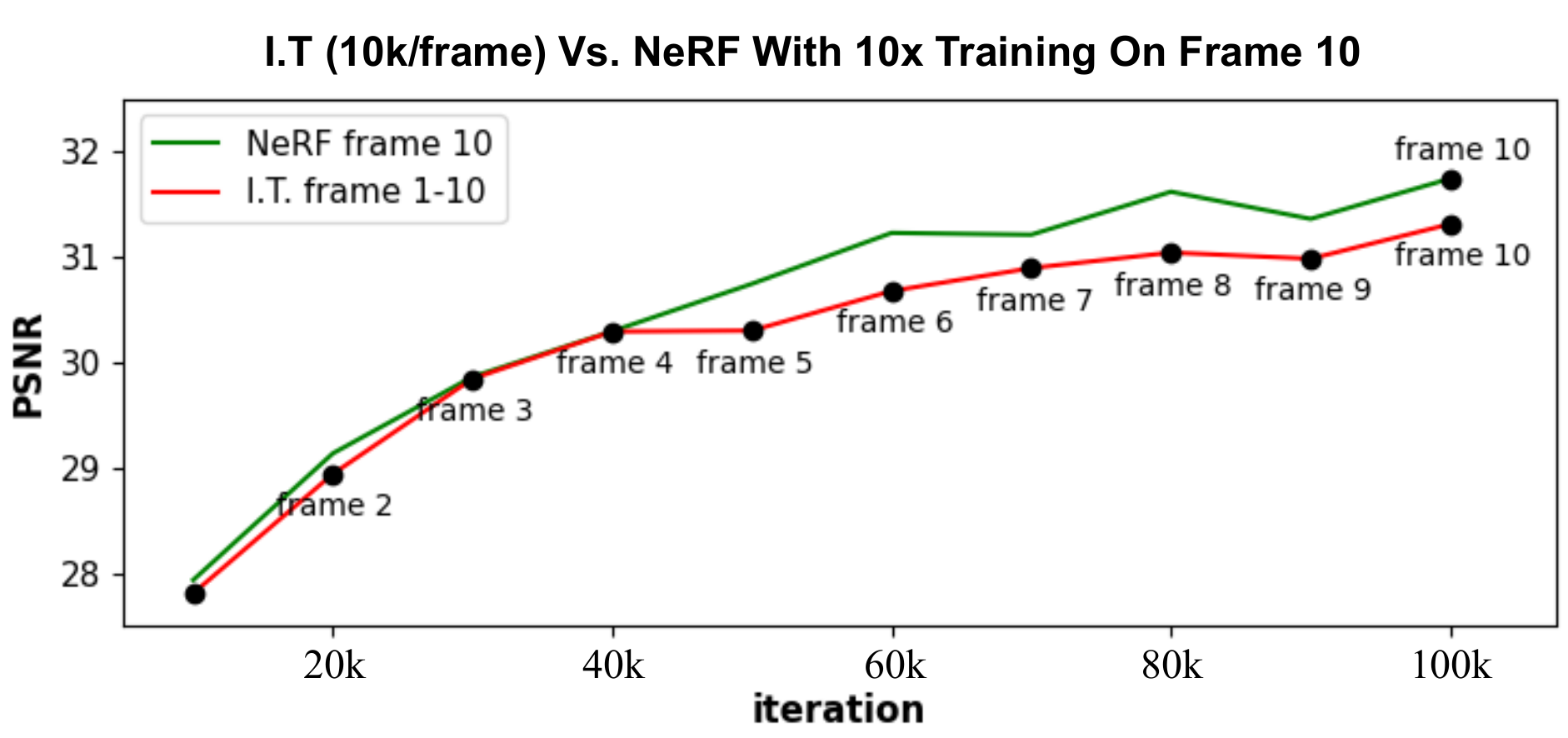}
   \caption{\textbf{NeRF 100k vs Incremental Transfer (I.T.) 10k:} On the "cut roasted steak" sequence in \textbf{\textit{Plenoptic Video Datasets}}\cite{neural3dvideos}, I.T. is trained for 10k iterations (8 min) for each of the 10 frames, thus accumulating the same 100k iterations as NeRF (from scratch). On frame 10, I.T. achieves 31.31db PSNR, and NeRF achieves 31.75db. I.T. continually improves upon previously learned information, and thus quickly generates good results with a fraction of the training.}
   \label{fig:nerf_vs_it}
\end{figure}
\setlength{\belowcaptionskip}{0pt}

We notice that, when the frame rate is infinitely high, adjacent frames should be almost identical. Therefore, given the MLP representation $\Phi_t$ for a frame at time $t$, it should take almost no effort to learn $\Phi_{t'}$ for the next frame at $t'$. Thus, the difference $\Delta\Phi=\Phi_{t'}-\Phi_{t}$ should decrease as time difference $\Delta t=t'-t$ decreases.
\[ \lim_{\Delta t \to 0} \Delta\Phi = \textbf{0} \]
Therefore, a model should reuse most of the previous information given high frame rates. Following this intuition, we propose Incremental Transfer (I.T.) to explicitly reuse information from frame to frame. We train one model per frame, with the earlier frame as the initialization for the later frame. As shown in Fig.\ref{fig:nerf_vs_it} and \ref{fig:overview}, I.T. leads to significantly less training by directly improving upon previous frames. The model thus gradually converges to a good quality similar to NeRF with only a fraction of the training. However, such a naive training scheme leads to significant temporal artifacts in rendered 3D videos, like flickering. In later sections, we show how INV reduces such artifacts. 

Moreover, in Table. \ref{table:quant}, we show that incrementally trained InstantNGP ("Ours NGP") acheives good results with merely 6 sec/frame of training.

% \begin{figure}
%   \centering
%   % \fbox{\rule{0pt}{2in} \rule{0.9\linewidth}{0pt}}
%    \includegraphics[scale=0.3]{images/color_scheme_transfer.png}
%    \caption{\textbf{Mixing structure and color/style in 2D:} We first train the MLP on image A for 2000 iterations. We then freeze all layers except the $1st$ layer, which is trained for 400 iterations on image B. The result is a mixture of A's color scheme and B's structure.}
%    \label{fig:color_scheme_transfer}
% \end{figure}

\subsection{Discovery of Structure and Color Layers}
\begin{figure*}
  \centering
   % \includegraphics[scale=0.45]{images/struct_swap.png}
  % \includegraphics[scale=0.21]{images/color_struct_info.png}
  \includegraphics[width=\textwidth]{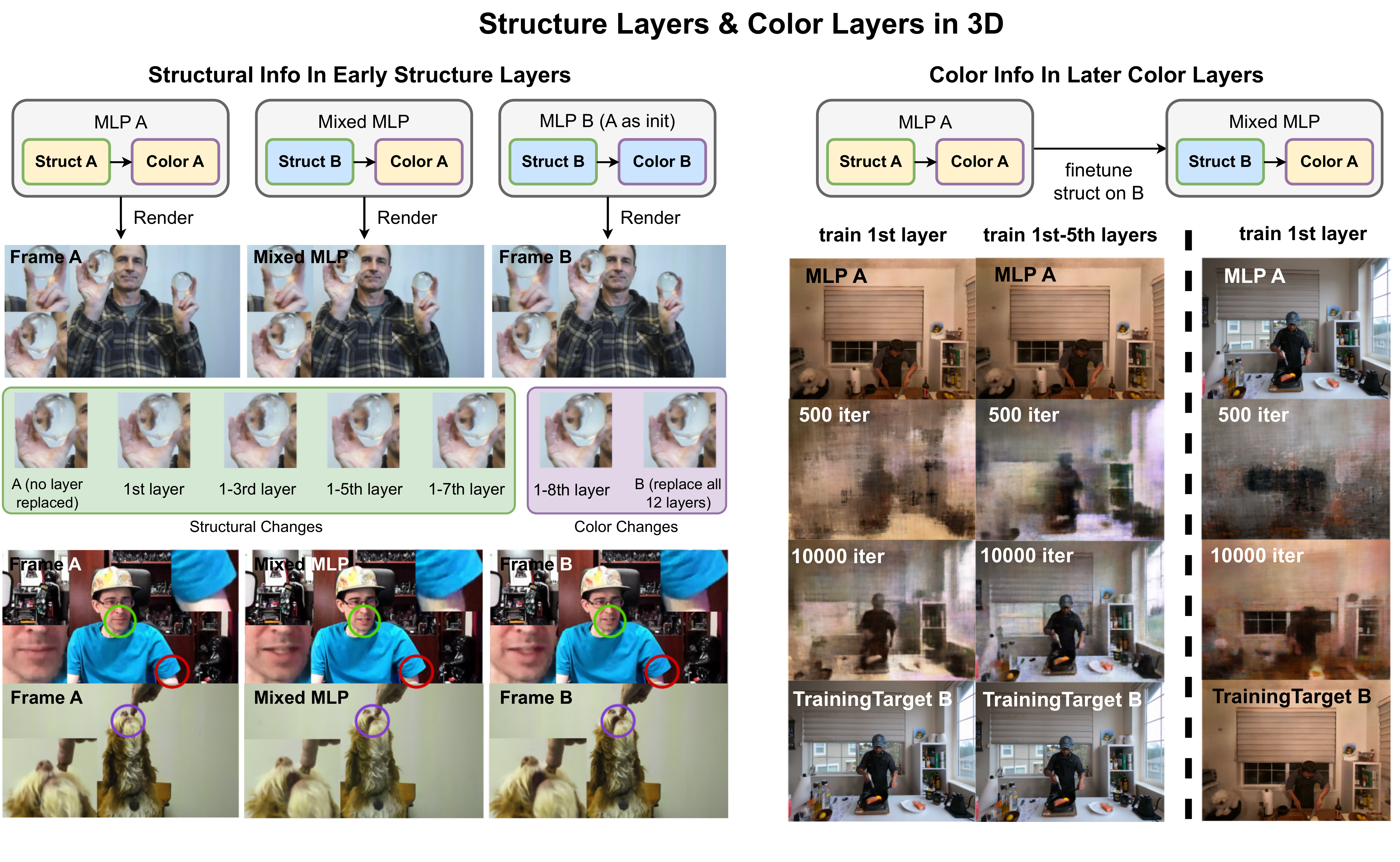}
  \vspace{-10mm}
   \caption{Natural information partitions also exist in NeRF. Different from 2D cases, there are more Structure Layers in the 3D case.
   }
   \label{fig:3d_swap}
\end{figure*}
Another benefit of Incremental Transfer is that we can now analyze the effect of different $\Delta\Phi$'s. This leads to our core discovery: When mapping from image/volume coordinates to colors (and/or densities), an MLP naturally partitions itself into Structure Layers and Color Layers. Structure Layers are early layers of the model that store structural information. Color Layers are later layers that store color/style information (\ie object color, shadows, etc.). Moreover, the mapping spaces learned by each layer are similar across nearby frames. We perform several experiments to show this interesting behavior.

\subsubsection{2D Structure Swap}
We first examine how changes in Structure Layers can affect the structural contents of 2D images. First, we perform Incremental Transfer to learn MLPs \textbf{(with positional encodings)} $\Phi_A$ and $\Phi_{B}$ for frames A and B. Second, we perform Structure Swap, where we replace the 1st layer of $\Phi_A$ with that of $\Phi_{B}$. As shown in Fig. \ref{fig:2d_swap} (Left), \textbf{without any further training}, the whole scene moves to the right (\eg more letters on the car plate are covered) and the kid's legs move closer. However, replacing later layers does not induce such structural changes. This means that 2D MLPs store structure information in earlier Structure Layers (the $1st$ layer in this case), and changes in Structure Layers induces structural changes in the rendered image. Moreover, since nearby Structure Layers can be swapped without further fine-tuning, the mappings learned by each layer are similar for nearby frames. This observation is important to support the idea of sharing the later layers.

\subsubsection{2D Color Scheme Transfer}
\label{sec:color_transfer}
We have shown that 2D MLPs \textbf{(with positional encodings)} store structural information in early Structure Layers. But what do later layers store? We explore the answer via the Color Scheme Transfer experiment. As shown in Fig. \ref{fig:2d_swap} (Right), we first train MLP $\Phi_A$ to reconstruct image A. Then, we finetune the Structure Layer (1st layer in this case) on image B while freezing later Layers. During the early stages of finetuning (\eg first 400 iters), the color scheme of A is often preserved while the structure quickly changes to B. Although the color scheme would eventually also change to B after longer training (\eg 1k iters), this phenomenon shows that color information is stored in the later Color Layers of a 2D MLP.

\subsubsection{3D Structure Swap and Color Scheme Transfer}
\label{sec:struct_swap}
We perform 3D analysis using our \textit{Immersive Telepresence Dataset} (Fig. \ref{fig:3d_swap} (Left)), captured with 3 camera bars, each containing 5 compact cameras. Our images are much closer to the subject, enabling much easier visualization and analysis. We observed the following behaviors:

(1) NeRFs often contain more than one Structure Layer (ranging from 3 to 7 layers depending on the content).

(2) Replacing incrementally more Structural Layers induces continual and meaningful changes/motions in 3D. For example in Fig. \ref{fig:3d_swap} (Left), when more Structure Layers from B are swapped into A, the person's head rotates more and a smile gradually appears. This means that the mappings learned by each layer stay consistent across time, allowing for layer replacements without finetuning.

(3) Replacing Colors Layers results in color changes (\eg tone, shadows, \etc) but no obvious structural changes. 

(4) 3D Color Scheme Transfer experiments show the same findings as 2D experiments.

This phenomenon is intuitively true since an MLP is a chain of non-linear mappings. The first layer maps coordinate/structural inputs to a latent space, which is then gradually transformed into the final color space through the chain of mappings. Earlier layers would thus retain more properties related to the input structural space.

3D Color Scheme Transfer experiments follow the same procedure described in Sec. \ref{sec:color_transfer}. In Fig. \ref{fig:3d_swap} (Right), we show that the novel view renderings demonstrate expected behaviors. In other words, during the early stages of training (\eg first 500 iters), the color scheme of the "original" is preserved while the structure quickly changes to the "target". Then, much longer training is needed to correct the color scheme since later layers are frozen (notice that with more trainable layers, this process becomes faster and easier). This consistency between 2D and 3D MLPs further supports our finding that MLPs \textbf{(with pos. enc.)} naturally store structure and color information separately.

These findings also hint at a different view of InstantNGP\cite{instantNGP}. \textbf{NGP's hash grid can be seen as the output of Structure Layers stored in 3D, and the MLP can be seen as the Color Layers}. In this way, NGP achieves the same "position $\rightarrow$ latent space" mapping as the Structure Layers without going through any linear layers, thus directly accessing the Color Layers and thus accelerating the training.

\begin{figure*}
  \centering
    % \fbox{\rule{0pt}{2in} \rule{.9\linewidth}{0pt}}
    \includegraphics[width=\textwidth]{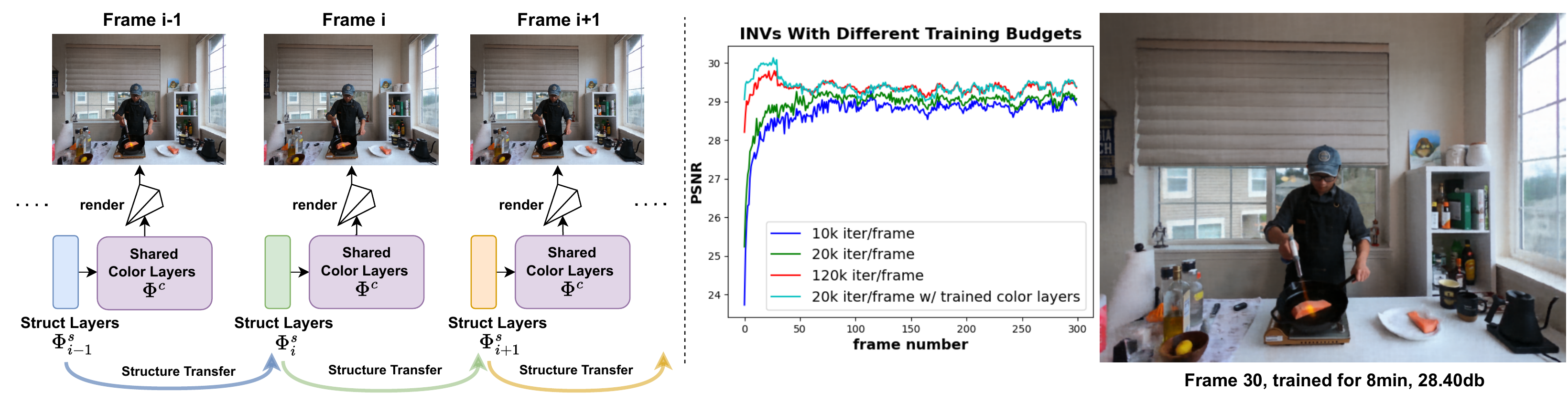}
    \caption{\textbf{Overview. Left:} Our representation consists of 2 types of modules: (1) Per-Frame Structure Layers (SL) $\Phi_i^s$ ($i\in {1...N}$), which encode the structure for a specific frame $i$, and (2) Shared Color Layers (SCL) $\Phi^c$, which are shared by all frames to describe the color style of the scene. A video of N frames results in N Per-Frame SLs $\Phi_i^s$ ($i\in {1...N}$) and 1 SCL $\Phi^c$. In this way, we maintain a constant color/style space while allowing the scene structure to change from frame to frame. \textbf{Right:} With only minutes of training per frame (8min in this example), INV quickly and continually improves in the first few seconds and then stabilizes to good quality.}
    \label{fig:overview}
\end{figure*}

\subsection{Efficient Incremental Neural Videos}
In Sec. \ref{sec:IT}, we showed that Incremental Transfer (I.T.) significantly reduces the training cost for the original NeRF while achieving good per-frame image quality. However, I.T. has two main disadvantages: (1) the storage cost is extremely high. A 5-minute 30FPS 3D video would require 40.1GB of storage. (2) There is significant flickering despite good per-frame image quality. To address these two issues, we propose the Incremental Neural Videos (INV) Representation that drastically reduces storage cost while providing high-quality visual results with notably improved temporal stability over I.T.
 % This is because both the color and structure are imperfectly fitted for each frame, and such imperfection is not consistent across frames. 

\subsubsection{The INV Representation}
% We notice that, since the frames of a coherent scene look similar to each other, the color space of the scene should change minimally.
Our representation consists of 2 types of modules: (1) Per-Frame Structure Layers (SLs) which are stored frame-by-frame, and (2) Shared Color Layers (SCL) which are shared by all frames. Per-Frame SLs encode the structure for a specific frame. They map the coordinate inputs (after positional encoding) to latent vectors $v$. SCL then maps these latent vectors $v$ into $RGB\sigma$. As a result, a video of N frames would be converted into N Per-Frame SLs $\Phi_i^s$ ($i\in {1...N}$) and 1 SCL $\Phi^c$. In this way, we have a constant color/style space for the scene while allowing the scene structure to change from frame to frame. Moreover, storing only Structure Layers brings storage savings. For the original NeRF of 12 layers (including $RGB$ and $\sigma$ heads, no skip connection), storing 3 Structure Layers would reduce the weight size to 1.12MB ($24.6\%$ of the complete model of 4.43MB). 

\subsubsection{Training INVs via Structure Transfer}
There are two stages to training INVs: 

(1) \textbf{Warm-Up}: Incremental Transfer (all layers trained) from frame to frame until frame $i$. Color Layers at frame $i$ are then stored and shared as the Shared Color Layers $\Phi^c$. 

(2) \textbf{Structure Transfer}: Incremental Transfer only on Structure Layers, with Shared Color Layers frozen. Structural information from previous frames is thus explicitly reused in later frames. Structure Layers' outputs are then converted into colors and densities via Shared Color Layers, which provide a constant color/style space. Only Structure Layers are stored for each frame (about 1.12MB/frame). Due to the freezing, videos are notably more stable as seen in the supplementary videos.

(Optional) \textbf{SplitINV: Video Stabilization with Background NeRF}: When INVs receive short training (\eg 8min/frame), videos can be flickery. To alleviate this, one can use a separate frozen NeRF to model the static backgrounds and focus computation on dynamic contents. Implementation details are included in the supplementary.

To render an INV frame $i$, one could first recover the full model $\Phi$ by concatenating the Per-Frame Structure Layers $\Phi_i^s$ with the Shared Color Layers $\Phi^c$. The recovered model $\Phi_i$ is then used to render the 3D frame $i$. 

\subsubsection{Temporal Weight Compression (TWC)}
To further reduce the size, we propose Temporal Weight Compression. Floating-point data compression algorithms (\eg \cite{fpzip,zfp}) can provide significant compression rate while maintaining high fidelity for structured data (\eg 2D images, 3D geometry, \etc), but they struggle to compress near-random data that lack clear structure, such as the MLPs weights. However, there could be temporal structures in how the weights change through time. Therefore, we construct temporal weight matrices by concatenating weights from consecutive frames. We then perform FPZIP\cite{fpzip} at 16bits precision on the temporal weight matrix. As shown in Table. \ref{table:quant} "INV+TWC", this simple approach maintains video quality while compressing the Structure Layers down to 0.3MB/frame. Therefore, once the Shared Color Layers (3.29MB) are transmitted, it only costs 0.3MB to transmit any new INV frame, \ie 72 MegaBits/second at 30FPS.

\begin{table*}[t]
  \centering
  \caption{\textbf{Comparisons and Ablations On Plenoptic Video Dataset\cite{neural3dvideos}} 
  With 50 minutes/frame less ($-19\%$) training than the SOTA Neural 3D Videos, INV achieves better per-frame qualities (PSNR $\&$ LPIPS).
  INV also achieves good quality in only 8min/frame. \\
  \textbf{"INV+TWC":} Temporal Weight Compression maintains video quality while compressing INV to 0.3MB/frame ($6.6\%$ of NeRF). \\
  % \textbf{"INV+Bgd":} With a separate frozen NeRF for static background, both quality and stability improves.
  \textbf{"INV$\star$":} with well-trained Shared Color Layers, it takes just 15min/frame to achieve similar quality to 90min/frame.
\textbf{Size$\downarrow$} and \textbf{Training$\downarrow$} are per-frame. \textbf{Warm-Up} is the number of warm-up frames, which are of lower quality but still included in the evaluation.
  }
  %For INV, we report the Per-Frame Structure Layer size (294KB/frame) since SCL only need to be transmitted/loaded once. 
  %We also show that, when InstantNGP\cite{instantNGP} is trained with Incremental Transfer (not sharing Color Layers), it can achieve state-of-the-art results in just 3 seconds.
  
   % \resizebox{0.48\textwidth}{!}{
\begin{tabular}{ l c c c c c c c c c}
\hline
 \multicolumn{1}{l}{\textbf{Method}} & \multicolumn{1}{l}{\textbf{Variation}} & \multicolumn{1}{l}{\textbf{Warm-Up}} & \multicolumn{1}{l}{\textbf{\# of SLs}} & \multicolumn{1}{l}{\textbf{PSNR$\uparrow$}} &  \multicolumn{1}{l}{\textbf{LPIPS$\downarrow$}} & \multicolumn{1}{l}{\textbf{JOD$\uparrow$}} & \multicolumn{1}{l}{\textbf{Size$\downarrow$}} & \multicolumn{1}{l}{\textbf{Training$\downarrow$}} \\
\hline
MVS & - & - & - & 19.1213  & 0.2599  & - & - & -\\

NV \cite{neuralvol} & - & - & - & 22.7975  &  0.2951  & 6.50 & \underline{2.58MB} & -\\

LLFF & - & - & - & 23.2388  &  0.2346  & 6.48 & - & -\\

NeRF-T & - & - & - &   \underline{28.4487}  &   \underline{0.1000}  & \underline{7.73} & - & -\\

DyNeRF\cite{neural3dvideos} & - & - & - &  \textbf{29.5808}  &  \textbf{0.0832}  & \textbf{8.07} & \textbf{0.09MB} & 260min\\

NeRF\cite{nerf} & - & - & - & 24.6232  & 0.3588 & 4.73 & 4.56MB & \textbf{8min}\\
%InstantNGP (15sec)\cite{instantNGP} \quad\quad& -   \quad\quad& -   \quad\quad& -  
InstantNGP\cite{instantNGP} & - & - & - &  24.4650 & 0.2977 & 4.87  & - & 6sec \\
\hline
Ours NGP & Incremental & 30 & - &  26.5933  & 0.0944 & 6.29  & - & 6sec \\
INV & - & 120 & first 3 layers &  28.7220  & 0.1215  & 6.61 & 1.12MB & \textbf{8min} \\
INV & - & 60 & first 3 layers &  28.9544  &  0.1111  & 6.79 & 1.12MB & \underline{15min}  \\
INV & - & 30 & first 3 layers &   \underline{29.3491}  &  \underline{0.0964} & \underline{7.08}  & 1.12MB & 90min\\
INV & - & 30 & first 3 layers &  \textbf{29.8471}  & \textbf{0.0818} & \textbf{7.42}  & 
1.12MB & 210min \\
\hline
INV & - & 60 & first 6 layers &  28.9236  &  0.1194  & 6.11 & 2.63MB & \textbf{8min} \\
INV+TWC & TWC & 120 & first 3 layers &  28.8698 & 0.1226 & 6.61 & \textbf{0.29MB} & \textbf{8min}\\
SplitINV & Bgd NeRF & 30 & first 3 layers & 28.7816 & 0.1157 & \underline{6.92} & 1.12MB & 10min\\
INV+TWC & TWC & 30 & first 3 layers &  \underline{29.3364} & \underline{0.1062} & \textbf{7.11} & \underline{0.31MB} & 90min\\
%INV+Bgd & Bgd Model & 30 & first 3 layers &  - & - & - & 1.04MB & \underline{9min}\\
INV \textbf{$\star$} & Trained SCL & 30 & first 3 layers &  \textbf{29.3791}  & \textbf{0.0898} & 6.85  & 1.12MB & \underline{15min} \\
\end{tabular}
\label{table:quant}
 % }
%\vspace{-2mm}
\end{table*}

\begin{table}[t]
  \centering
  \caption{\textbf{INV Evaluation on other sequences}: We evaluate INV on 3 other sequences released by Neural 3D Videos(DyNeRF) \cite{neural3dvideos}. We are not able to compare to DyNeRF on these sequences because DyNeRF did not report the performances on these sequences and the code is not released.} 
  \resizebox{0.48\textwidth}{!}{
\begin{tabular}{ l c c c c c c c c c}
\hline
 \multicolumn{1}{l}{\textbf{Seq}} & \multicolumn{1}{l}{\textbf{PSNR$\uparrow$}} &  \multicolumn{1}{l}{\textbf{LPIPS$\downarrow$}} & \multicolumn{1}{l}{\textbf{JOD$\uparrow$}} & \multicolumn{1}{l}{\textbf{Training$\downarrow$}} \\
\hline

cut roasted beef & 32.0214  & 0.0644  & 7.17 & 20k (15min) \\
sear steak & 31.7662 & 0.0518 & 7.51 & 120k (90min) \\
coffee martini & 28.1384 & 0.0963 & 6.90 & 120k (90min) \\
\end{tabular}
\label{table:other_quant}
}
\end{table}

\section{Implementation Details}
\label{sec:implementation}
Our baseline NeRF is a faithful PyTorch re-implementation of NeRF that reproduces the original results. We use the same NeRF for INV, except that we disable skip connections to the middle of the MLP. Although skip connections provide some quality improvement, they activate more Structure Layers and thus increase the storage size. All models are trained on a single NVIDIA RTX3090 GPU with the same settings as the original NeRF (learning rate $5\times10^{-4}$, 1024 ray samples, 64 depth samples for coarse NeRF and 128 samples for fine NeRF). In our tables, 8 minutes of INV training amounts to 10k iterations, 15 min is 20k, 90 min is 120k, and 210 min is 280k.

Temporal Weight Compression batches all INV frames (after warmup) into a single temporal weight matrix, then we use the \textit{fpzip} python library to perform floating-point compression at 16 bits resolution.

\section{Experiments}
We evaluate the quantitative and qualitative performance of INV against State-Of-The-Art approaches. 

\begin{figure*}
  \centering
  \includegraphics[width=\textwidth]{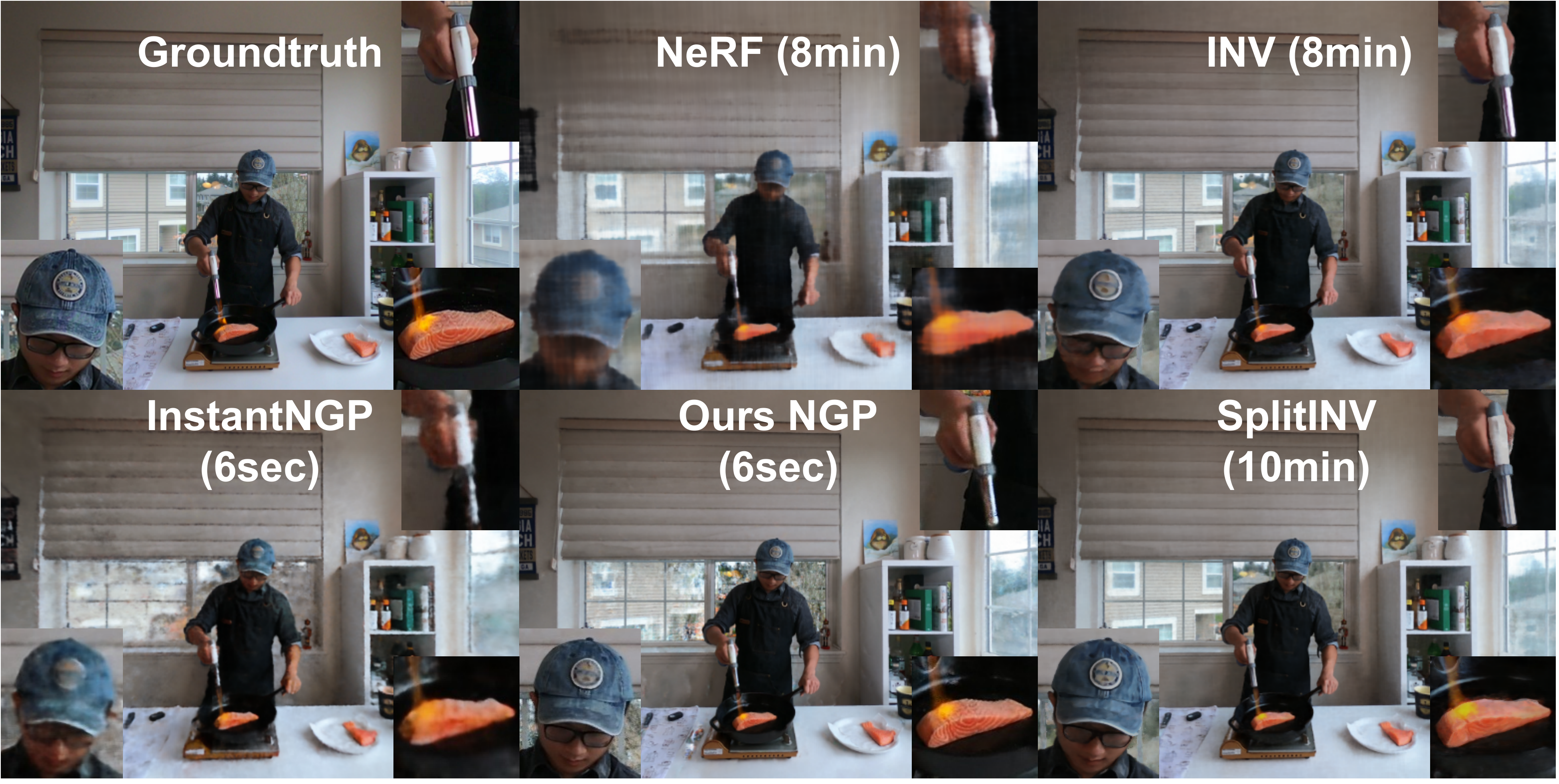}
   \caption{\textbf{Visual Comparisons}: INV achieves significantly better results than NeRF training just 8min/frame, and SplitINV renders even better foreground with just 10min/frame. "Ours NGP" is trained incrementally and achieves significantly better results than the original InstantNGP. Notice that "Ours NGP" achieves good LPIPS score but low PSNR score because it achieves high qualities on large portions of the scene, but often suffers from large amounts of floating artifacts (\eg windows, around the person)
   }
   \label{fig:comparison}
\end{figure*}

\subsection{Datasets}
The \textbf{\textit{Plenoptic Video Datasets (PVD) \cite{neural3dvideos}}} is released by Neural 3D Videos (DyNeRF) as their benchmark dataset. It is captured with 21 GoPro cameras at $2028\times2704$ and 30 FPS. It contains challenging volumetric effects (\eg flames), view-dependent effects (e.g. specularity from silverware, transparency from bottles), complex actions (\eg cooking), changing topologies (\eg cutting beef, pouring liquid), etc. Same as DyNeRF, INV is evaluated at $1352\times1014$ on the same and only sequence that DyNeRF reported quantitative results on, \ie the first 10 seconds of the "flame salmon 1" sequence. We report quantitative results on other sequences in the supplementary.

\textbf{\textit{Immersive Telepresence Dataset}}. As mentioned in Sec. \ref{sec:struct_swap}, we also perform analysis of Structure and Color Layers on our custom dataset. The dataset is captured with 3 camera bars each containing 5 compact cameras, and we follow \textit{PVD \cite{neural3dvideos}} to use 14 views for training and 1 middle view for evaluation. Different from \textit{PVD \cite{neural3dvideos}}, our dataset is captured closer to the subject and thus much more suitable for analyzing and visualizing Structure Swap. However, since compact cameras have lower image qualities, this dataset is only suitable for visualization and analysis but quantitative evaluation. Please refer to the supplementary for more details.

\subsection{Metics and Competing Methods}
We follow DyNeRF to use these metrics:(1) Peak Signal-To-Noise Ratio (PSNR), (2) Perceptual Score (LPIPS\cite{LPIPS}), and (3) Just-Objectionable-Difference(JOD)\cite{JOD}. We omit FLIP and DSSIM since we deem the three widely used metrics enough for reliable measurement. The following methods are evaluated:

\textbf{NeRF \cite{nerf}}: Our baseline NeRF is trained from scratch and stored for each frame separately. This baseline helps to quantify the training acceleration and storage savings provided by INV (which uses the same NeRF internally).

\textbf{Neural 3D Videos (DyNeRF) \cite{neural3dvideos}}: DyNeRF proposed to use learnable latent codes as inputs for each frame (in addition to positional encoding) to improve the rendering quality. DyNeRF also accelerates training via hierarchical training and ray importance sampling. Longer videos are divided into chunks of 10-second clips for better results. The authors also released the \textit{Plenoptic Video Datasets \cite{neural3dvideos}}. They show showed superior performance to their baselines. Since the \textbf{DyNeRF code is not available} and only results on the "flame salmon" sequence are reported, we cannot provide side-by-side visual comparisons and can only compare with them on that sequence.

\textbf{InstantNGP \cite{instantNGP}}: We train InstantNGP on the scene with 800 iterations (\~6sec) from scratch for each frame.

%%%%%%%%%    Where Should The NGP Stuff Go?

% \textbf{Per-Frame Instant NGP \cite{instantnerf}: Similar to Per-Frame NeRF, the original Instant NGP model is trained from scratch for each of the frames separately. This baseline helps to evaluate the effectiveness of Incremental Transfer on Instant NGP, where an explicit grid is used to represent the structure instead of MLPs.

% \textbf{NeRF + Incremental Transfer}: The original NeRF model is trained with Incremental Transfer, but without freezing/sharing the color layers.

% \textbf{Instant NGP + Incremental Transfer}: The original NeRF model is trained with Incremental Transfer, but without freezing/sharing the color layers.

\textbf{INV}: INV uses the original NeRF with the same settings as our NeRF baseline, but we disable skip connections. This is because they activate more Structure Layers, leading to higher storage costs while providing insignificant improvements. In most experiments, we train 3 Structure Layers. Although the frozen and shared layers also include some Structure Layers, we count them as part of the Shared Color Layers for simplicity. "INV+TWC" uses TWC to batches and compress all frames after Warm-Up. "SplitINV" uses a separate frozen NeRF to model the backgrounds, and INV is used to capture the foreground. All metrics are measured on all 300 frames, including the Warm-Up frames which show worse results. 

\textbf{SplitINV}: We use a separate NeRF to model the static background, while the foreground is encoded by INV. This allows INV to focus computation on the foreground and also helps stabilize the background. Please refer to the supplementary for more details on training. 

\textbf{Ours NGP}: We train InstantNGP on the scene with 800 iterations/frame with the previous frame as the initialization for the MLP and hash grid. Importantly, \textbf{the hash grid of NGP can be seen as the output of Structure Layers stored in 3D, and the MLP can be seen as the Color layers}. In this way, NGP skips the Structure Layers and directly accesses the Color layers. Therefore, after 30 frames of warmup, we fix the MLP and train the hash grid only on the foreground pixels (as estimated by an optical flow estimator\cite{separableflow}). This allows NGP to focus computation on the foreground and also helps stabilize the background.

\textbf{Multi View Stereo (MVS), Local Light Field Fusion (LLFF), NeuralVolumes (NV), and NeRF-T}: We adopt the numbers as reported by DyNeRF\cite{neural3dvideos}. Please refer to \cite{neural3dvideos} for more details.

\begin{figure*}
  \centering
  \includegraphics[width=\textwidth]{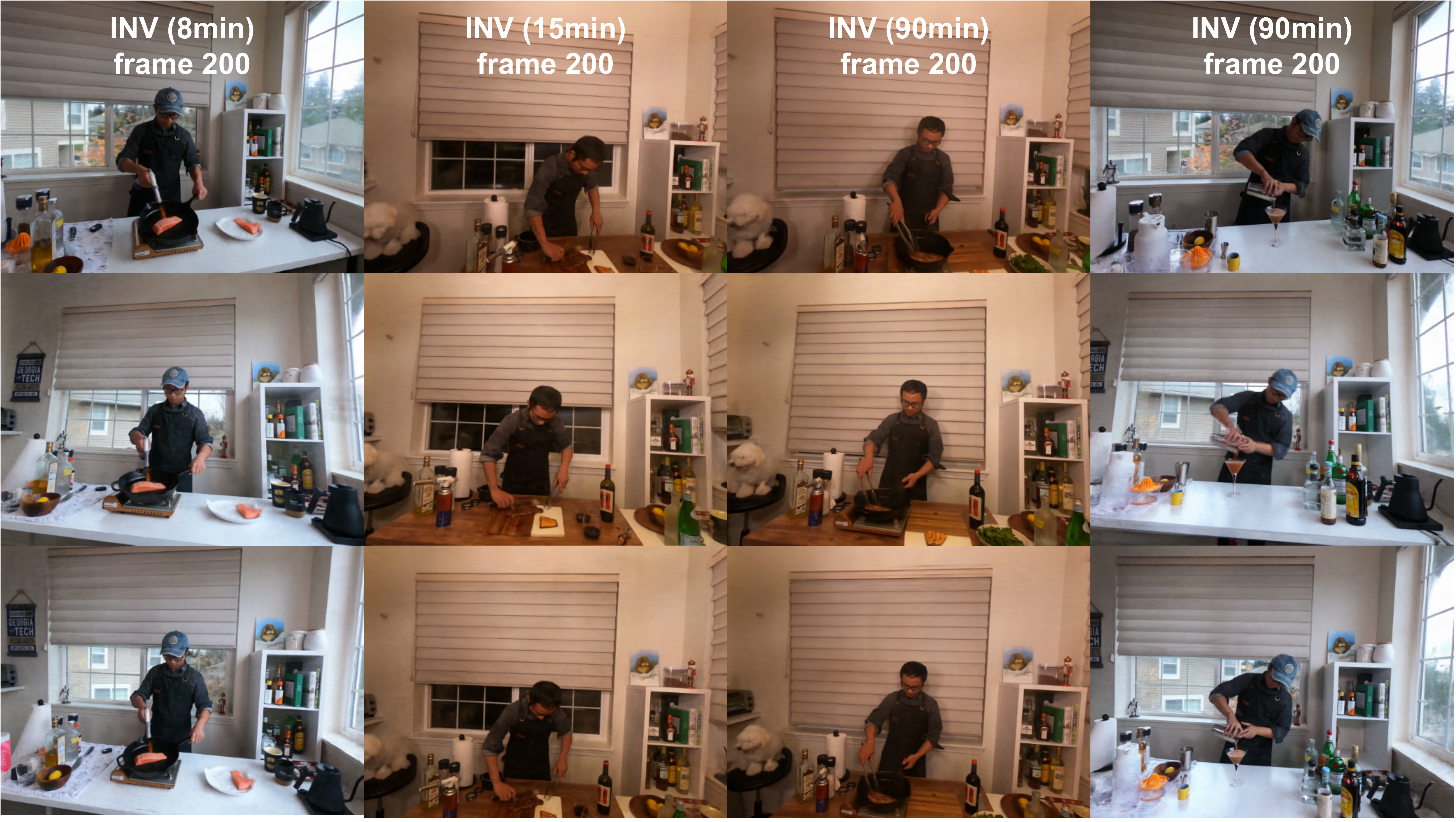}
   \caption{\textbf{Free-Viewpoint Renderings}: We show free viewpoint renderings on the 200th frames of sequences "flame salmon 1", "cut roasted beef", "sear steak", and "coffee martini".
   }
   \label{fig:comparison}
\end{figure*}

\subsection{Results}
As shown in Fig. \ref{fig:nerf_vs_it} and \ref{fig:overview}, INV's quality quickly and continually improves during the Warm-Up Stage. During the Structure Transfer Stage, INV performance stabilizes at a good quality. Moreover, INV can model challenging visual effects (\eg flames, reflections, \etc) even though the Color Layers are frozen and shared between frames. 

As shown in Table. \ref{table:quant}, INV achieves state-of-the-art results on per-frame quality metrics (PSNR \& LPIPS) with 50 minutes/frame less ($-19\%$) training than DyNeRF. Moreover, with merely 8min/frame of training, we achieve an average PSNR of 28.77db ("INV+TWC"). We also notice that more trainable Structure Layers result in better per-frame image metrics, but it almost completely removes the stabilization effect of freezing later layers as seen in the table and supplementary videos. This is because more structural contents are allowed to change with no guarantee that the change is consistent over time.

In "INV$\star$", we show that well-trained Shared Color Layers can notably improve quality. During Warm-Up, the models are trained for 210min/frame. During Structural Transfer, the Structure Layers are only trained for 15min/frame. However, compared to the basic INV trained for 15min/frame throughout the video, PSNR increases by 0.42db (from 28.95db to 29.38db). This quality is similar to the basic INV trained for 90min/frame. Pretraining INV could potentially be useful for many live-streaming settings (\eg game streaming, lectures, and remote meetings). Since many live-streaming sessions capture the same background and people, it could be helpful to maintain or continually improve a long-term memory via a color/style space that is shared across sessions. Moreover, "SplitINV" increases the stability via its frozen background NeRF.

Due to its incremental nature, INV can synthesize 3D videos on-the-fly and does not have to wait for video chunks before processing. This property makes it naturally suitable for interactive 3D streaming. Additionally, INV doesn't make strong assumptions about the modeling backbone, thus it can likely be enhanced by more sophisticated techniques.

\section{Limitations and Future Work}
The main limitation of our work is visual stability. INV models that are trained longer (\eg 210min/frame) generally show good stability. However, short training (\eg 8min/frame) results in poor stability despite good per-frame qualities. While background models could improve stability on static contents, it is still challenging to stabilize dynamic contents effectively. It is likely helpful to enforce consistency for temporal correspondences (\eg as done in \cite{neural3dvideos,nsff,dynamicnerf}). Moreover, it could improve stability and quality by training an intelligently selected subset of neurons, instead of always freezing certain layers. We leave this for future research.

While INV shows promising training acceleration, more needs to be done to achieve real-time training in the future. One advantage of INV is that it retains and continually improves upon previous information. This property could be combined with other approaches (\eg smarter sampling, human-specific modeling, voxel-based approaches, etc.) to further reduce training time.

Additionally, we did not address the need for real-time rendering in this work. Since there are many prior works \cite{fastnerf,bakingnerf,plenoctrees,instantNGP} on real-time rendering for NeRFs, one might be able to achieve real-time streaming and rendering by combining INV with these approaches, assuming that the INV frames are already trained.

Another future work is to handle large and/or abrupt changes, as well as moving cameras. A simple solution is to detect these sudden changes and update the entire model. It could also be beneficial to dynamically determine which layers or neurons to train. However, this is likely less of a problem for many live-streaming scenarios with still cameras and smoothly varying content (\eg game streaming, lectures, \etc). For general videos, it is also possible to store dedicated Shared Color Layers for each scene.

% (2) While we showed a significant reduction in storage sizes, there is still a lot to be done to enable streaming usage. Currently, Per-Frame Structure Layers consume about 300 Kilo-Bytes/frame (about 70 Mega-Bits/second), which is easily attainable by institutional internet (\eg universities, businesses, \etc) but could be overwhelming for typical household internet bandwidth (US national average)

  % \item Unlike approaches like NSFF\cite{nsff}, approaches like ours and N3V\cite{neural3dvideos} do not generate scene flow as a byproduct. Sceneflow might be important for related tasks like action recognition, segmentation, \etc.
  
\section{Conclusions}
In this work, we introduced INV, an efficient frame-by-frame representation naturally suitable for interactive 3D streaming. It significantly reduces training while compressing storage sizes to the streamable realm. Additionally, we find that MLPs naturally partition themselves into Structure and Color Layers. Our findings can thus help the community better understand and manipulate MLP models. We believe that this is a solid step toward the future of interactive 3D streaming. 

{\small
\bibliographystyle{ieee_fullname}
\bibliography{egbib}
}

\end{document}